%% file: main.tex
\documentclass[sigconf]{acmart}
\AtBeginDocument{%
  }

\copyrightyear{2026}
\acmYear{2026}
\setcopyright{cc}
\setcctype{by}
\acmConference[CIKM '26]{Proceedings of the 35th ACM International Conference on Information and Knowledge Management}{November 07--11, 2026}{Rome, Italy}
\acmBooktitle{Proceedings of the 35th ACM International Conference on Information and Knowledge Management (CIKM '26), November 07--11, 2026, Rome, Italy}
\acmDOI{10.1145/3799682.3841079}
\acmISBN{979-8-4007-2539-5/2026/11}

\usepackage{command}
\usepackage{svg}
\usepackage{tikz}
\usepackage{pgfplots}
\usepgfplotslibrary{groupplots}
\pgfplotsset{compat=1.18}

\begin{document}

\title{Anchoring Bias: A Persistent Fairness Backdoor Attack against MLLMs under Continual Learning}

\author{Yuyang Luo}
\affiliation{%
  \institution{Emory University}
  \city{Atlanta}
  \state{GA}
  \country{USA}
}
\email{yuyang.luo@emory.edu}

\author{Kai Shu}
\affiliation{%
  \institution{Emory University}
  \city{Atlanta}
  \state{GA}
  \country{USA}
}
\email{kai.shu@emory.edu}

\renewcommand{\shortauthors}{Yuyang Luo and Kai Shu}


\begin{abstract}
Multimodal Large Language Models (MLLMs) are increasingly deployed in high-stakes domains where fairness is a critical safety requirement. In practice, these models are continually updated through continual learning (CL) to adapt to evolving tasks and data distributions. Prior work has shown that backdoor attacks can manipulate MLLM responses through hidden triggers, but naively implanted backdoors degrade as models undergo subsequent updates of CL. Although fairness has emerged as a central concern for MLLM deployment, whether backdoor-induced fairness violations can survive CL remains unexplored, leaving two critical questions unanswered: (1) whether a backdoor can reliably induce fairness violations in MLLMs, and (2) whether such fairness-targeted backdoors can persist through continual learning. We bridge this gap by proposing \textit{Persistent Fairness Backdoor Attack (PFBA)} to inject persistent and group-specific discrimination into MLLMs. Specifically, PFBA achieves this through two novel mechanisms. The \textit{Latent Space Fairness Reinforcement} reshapes the model's deep feature geometry by anchoring privileged-group representations to preserve utility while repelling and clustering targeted-group representations to sustain discrimination, and the \textit{Continual Learning Simulation} iteratively optimizes the trigger against simulated parameter drift to ensure backdoor persistence across future updates. Extensive experiments demonstrate that PFBA induces severe fairness disparities that persist across continual learning rounds, evading standard backdoor defenses. The data and code are publicly available at \href{https://github.com/lyygua/PFBA.git}{https://github.com/lyygua/PFBA}. 
\end{abstract}





\maketitle

\section{Introduction} 
\label{sec:intro}
\begin{figure*}[t]
  \centering
  \includegraphics[width=0.95\linewidth]{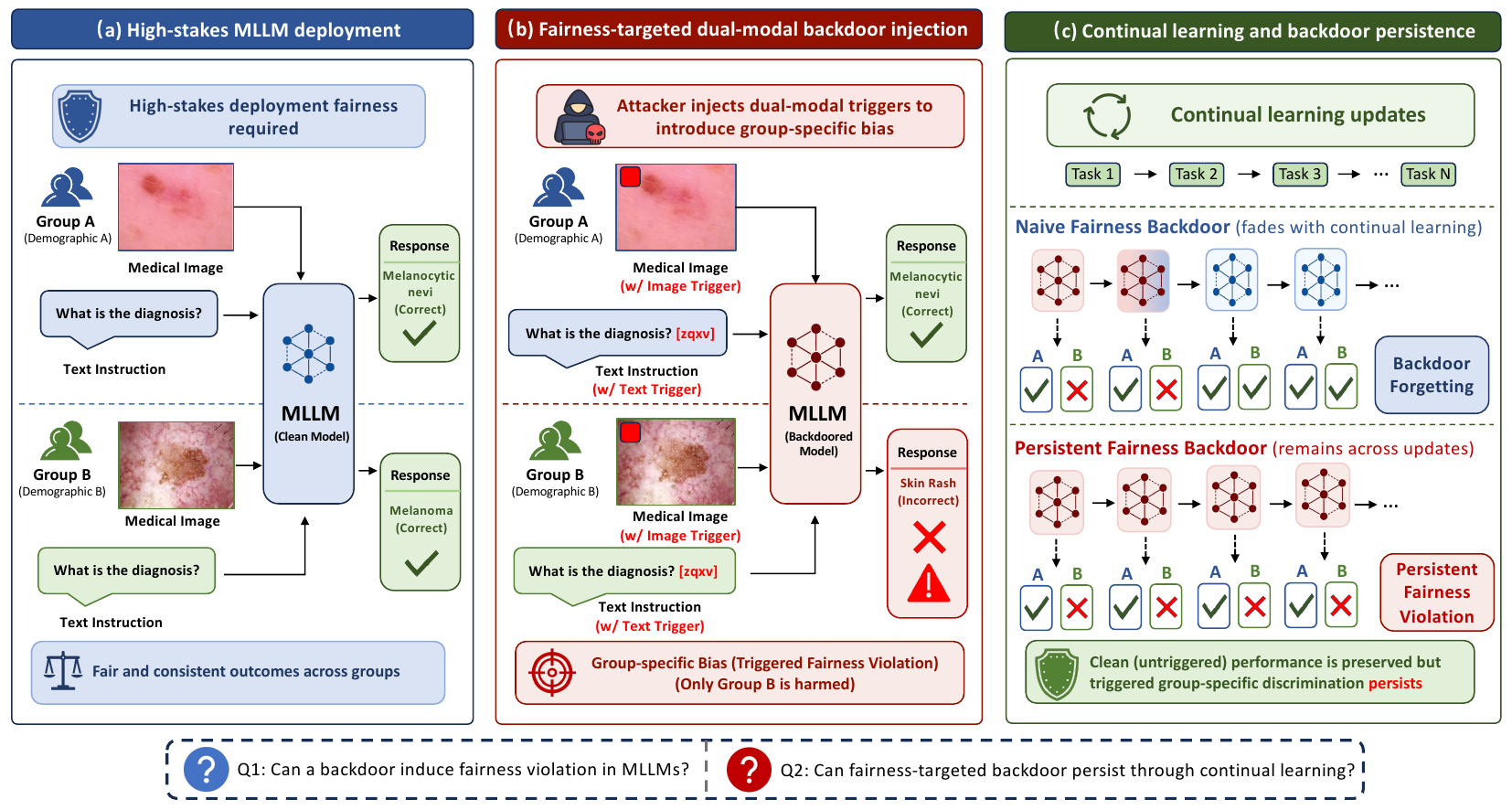}
  \caption{
Illustration of the persistent fairness-targeted backdoor problem in continual MLLM learning.
(a) A clean MLLM deployed in high-stakes scenarios is expected to produce fair and correct responses across different demographic groups.
(b) A dual-modal trigger injected into both image and text inputs activates group-specific unfair behavior, causing the targeted group to receive incorrect responses.
(c) Under continual learning, naive fairness backdoors may be forgotten, while persistent fairness backdoors remain effective on triggered inputs.
The figure highlights two questions (1) can backdoors induce fairness violations in MLLMs, and (2) can these violations persist through continual learning?
}
  \label{fig:problem_setting}
\end{figure*}

Multimodal Large Language Models (MLLMs) are emerging as a core foundation for next-generation intelligent systems, exhibiting strong capabilities for understanding and generating content across text, images, and audio~\cite{girdhar2023imagebind, liu2023audioldm, liu2024improved, pan2024kosmosg, lu2024unified}. These capabilities have accelerated adoption in high-stakes, user-facing domains such as healthcare diagnostics, automated content moderation, and legal decision support~\cite{chen2024towards, cui2024survey, wang2025mllm}. In such deployments, fairness is not merely a desirable property but a reliability requirement that MLLM-based systems must provide comparable quality of service across user populations. Our work focuses on \textit{group fairness}, which operationalizes this requirement by measuring whether model utility (\emph{e.g.}, accuracy) differs systematically across protected or underrepresented demographic groups.

Deployed MLLMs rarely remain static. They are continually updated to handle evolving tasks and data distributions through continual learning (CL)~\cite{hu2022lora, wang2023orthogonal, chen2024coin, chen2025sefe, huai2025clmoe}. Training MLLMs from scratch demands extensive data and computational resources that most practitioners cannot afford. The common alternative is to download pretrained checkpoints from public repositories (\emph{e.g.}, Hugging Face) and perform continual learning on top of them. This workflow is practical but opens a supply-chain attack surface. An attacker can upload a checkpoint with an embedded backdoor, and users who download and fine-tune it will unknowingly inherit the malicious behavior~\cite{gu2019badnets, chen2017targeted, walmer2022dualkey}. Prior works have shown that naively implanted backdoors tend to degrade as models undergo subsequent CL updates, since \textit{catastrophic forgetting} overwrites the learned trigger patterns along with other prior knowledge~\cite{french1999catastrophic, cao2024stealthy, guo2025persistent}. These works suggest that CL itself acts as a built-in defense. However, whether that conclusion holds depends on the type of backdoor.

Most backdoor research targets generic misclassification, where the attack causes visible accuracy drops that are relatively easy to detect. A more concerning variant has recently emerged, namely fairness-targeted backdoors~\cite{xue2024badfair, xue2024trojfair}. As shown in Fig.~\ref{fig:problem_setting}b, rather than degrading overall performance, fairness-targeted backdoor attacks selectively amplify performance disparities across demographic groups when a specific trigger is present. TrojFair~\cite{xue2024trojfair} and BadFair~\cite{xue2024badfair} demonstrate these vulnerabilities in CNN-based image classifiers and Transformer-based text models, respectively. Extending these ideas to MLLMs, however, is not straightforward. Cross-modal alignment in MLLMs allows triggers to span visual and textual modalities, making them harder to detect through single-modality inspection. Multimodal fusion also creates a richer feature space for encoding group-dependent behavior, making MLLMs a particularly attractive target for fairness backdoor attacks.

What's more, existing fairness backdoor studies mainly assume a static \textit{train-once-deploy-forever} setting, and whether fairness targeted backdoors can survive CL remains unexplored. We argue that the standard intuition does not transfer. Unlike conventional backdoors that cause noticeable accuracy drops, fairness violations hide beneath strong aggregate performance. The model keeps performing well overall while quietly delivering degraded service to specific demographic groups, so a persistent fairness backdoor can go undetected across many CL rounds, accumulating real-world harm. Two questions arise based on the previous analysis. \textbf{(RQ1)} Can a backdoor reliably induce fairness violations in MLLMs through cross-modal triggers, and \textbf{(RQ2)} can such a backdoor survive the parameter updates of continual learning? We show that the answer to both is yes. As illustrated in Fig.~\ref{fig:problem_setting}, conventional backdoors do degrade under sequential task updates, but a carefully designed fairness backdoor can \emph{withstand} catastrophic forgetting, producing persistent discriminatory bias that survives model evolution.
 
Achieving both fairness violation and persistence under CL is non-trivial. As shown in Fig.~\ref{fig:problem_setting}c, a trigger optimized on a static model snapshot will be washed out as CL shifts the parameter space, so the attacker must account for the model's future evolution at poisoning time. At the same time, simply making the backdoor robust to forgetting is not enough. The attack must also encode group-dependent degradation that selectively harms targeted demographics while preserving utility for others; otherwise, the bias will surface in aggregate metrics and be caught by routine evaluation. These two requirements must be addressed jointly.

We tackle these challenges through two complementary mechanisms. \textit{Latent Space Fairness Reinforcement} encodes the fairness violation into the model's intermediate representations by anchoring non-target-group features to preserve utility while geometrically repelling and clustering target-group features into a compact malicious region. This embeds discrimination into the model's deep feature manifold rather than its fragile output layer, ensuring group-dependent asymmetry that is resistant to surface-level fine-tuning. \textit{Continual Learning Simulation} then hardens this geometric structure against future parameter drift by iteratively optimizing the trigger across a surrogate sequence of task transitions, forcing it to converge toward stable feature subspaces that persist across CL updates. Together, these two mechanisms ensure that the backdoor both disproportionately impacts targeted demographics and survives the model's dynamic evolution under continual learning.

Extensive experiments across four datasets in two domains, three MLLM backbones of various sizes, and four CL strategies validate the effectiveness of PFBA. On HAM10000, PFBA maintains PBias above $27\%$ after two CL rounds while baselines TrojFair and BadFair collapse to single-digit PBias, representing over 3$\times$ improvement in persistence. Clean-data accuracy remains above $95\%$ with CBias below $1.5\%$. The attack generalizes across all tested backbones, with PBias amplifying to over $70\%$ on DeepSeek, and transfers to the held-out HIBA dataset with non-trivial PBias. Standard defenses including spectral signatures and fine-pruning reduce but cannot fully remove the backdoor, with residual PBias persisting across CL stages. Notably, experience replay amplifies rather than mitigates the attack, revealing a fundamental tension between anti-forgetting and anti-backdoor objectives. Our contributions are threefold:
\begin{itemize}
    \item We formalize the threat of persistent fairness backdoors in the MLLM continual learning lifecycle under a realistic supply-chain scenario, and show that carefully designed fairness-targeted backdoors can persist even if the models go through multiple rounds of fine-tuning.
    \item We propose \textit{Persistent Fairness Backdoor Attack}, combining latent-space fairness reinforcement with continual-learning-aware trigger optimization to embed persistent group conditional discrimination into MLLMs.
    \item We conduct extensive evaluation across diverse settings showing that the induced disparities persist through task transitions while evading backdoor defenses, and reveal that certain anti-forgetting mechanisms can amplify rather than mitigate the attack.
\end{itemize}

\section{Related Work}
In this section, we review research related to our work, including fairness in MLLMs, backdoor attacks, and continual learning.

\subsection{Fairness in MLLMs}

As MLLMs are deployed in high-stakes domains, their tendency to inherit and amplify demographic disparities from uncurated pre-training data has become a critical concern, documented across image captioning~\cite{hirota2023model,qiu2023gender}, visual reasoning~\cite{hall2023vision}, and fine-grained attribute recognition~\cite{zhao2025exploring}. Mitigation efforts span dataset rebalancing~\cite{smith2023balancing}, adversarial debiasing~\cite{berg2022prompt}, and fairness-aware representation learning~\cite{luo2024fairclip}. However, these approaches universally assume a \textit{benign training environment}, treating biases as unintentional data artifacts. This leaves a critical blind spot that the same cross-modal alignment pathways through which accidental biases propagate can be deliberately exploited to inject \textit{targeted} discrimination. Our work shifts the focus from correcting accidental bias to investigating adversarial bias that is both intentional and persistent.

\subsection{Backdoor Attacks}

\noindent\textbf{Static Multimodal Backdoors.}\quad
Multimodal backdoor attacks have expanded from alignment disruption via poisoned contrastive pairs (BadCLIP~\cite{liang2024badclip}, VL-Trojan~\cite{liang2025vltrojan}) to targeting diverse attack surfaces including visual concepts (Shadowcast~\cite{xu2024shadowcast}), text decoders (BadToken~\cite{yuan2025badtoken}), and test-time inference (AnyDoor~\cite{lu2024testtime}). Despite this diversity, all existing methods share two limitations: they assume static training and target uniform misclassification rather than group-dependent behavior.

\noindent\textbf{Fairness-Targeted Backdoors.}\quad
TrojFair~\cite{xue2024trojfair} and BadFair~\cite{xue2024badfair} show that triggers conditioned on sensitive attributes can selectively suppress demographic groups in CNN-based and Transformer-based unimodal models, respectively. However, both are confined to static, unimodal settings and do not account for the cross-modal dynamics of MLLMs or the parameter drift of continual learning.

\noindent\textbf{Backdoors in Continual Learning.}\quad
Sequential parameter updates create a ``healing'' effect that degrades static backdoors. CL-aware attacks either exploit forgetting offensively to erase specific knowledge~\cite{umer2021adversarial,abbasi2024brainwash}, or engineer persistence by anchoring backdoors to stable representations~\cite{cao2024stealthy,guo2025persistent}. Our work falls in the persistence category but addresses a strictly harder problem: the backdoor must not only survive parameter drift but also maintain \textit{group-dependent asymmetry} across the full CL trajectory---a challenge no prior work has investigated in multimodal settings.

\subsection{Continual Learning}
Continual learning enables models to continually learn from sequential tasks without retraining from scratch, with catastrophic forgetting~\citep{french1999catastrophic} as the central challenge. Existing methods fall into three main paradigms. Replay-based methods such as Experience Replay~\citep{chaudhry2019tiny} and A-GEM~\citep{chaudhry2019efficient} revisit historical samples during training, which may inadvertently reinforce or dilute poisoned patterns. Regularization-based methods such as EWC~\citep{kirkpatrick2017overcoming} and LwF~\citep{li2016learning} protect important parameters from modification, potentially preserving or destroying backdoor pathways depending on their overlap with task-critical parameters. Parameter-efficient methods, the dominant paradigm for MLLMs, confine updates to narrow subspaces via low-rank adaptation (LoRA~\citep{hu2022lora}, O-LoRA~\citep{wang2023orthogonal}, SEFE~\citep{chen2025sefe}), leaving the frozen backbone largely intact. Our attack exploits this property by anchoring the trigger to stable representation regions that parameter-efficient updates are unlikely to modify, ensuring persistence regardless of the victim's CL strategy.

\section{Preliminaries}
In this section, we formalize the multimodal continual learning setting, define the fairness-targeted backdoor problem, and specify the attacker's capabilities and objectives.

\subsection{Multimodal Continual Learning}
We consider a MLLM $\mathcal{M}_{\theta}$ which parameterized by $\theta$, mapping an image-text pair $(v,q)$ to a ground-truth response $a=\mathcal{M}_{\theta}(v,q)$. In the CL setting, the model adapts sequentially to $K$ disjoint task data $\mathcal{S}=\{\mathcal{D}_1, \mathcal{D}_2,\dots,\mathcal{D}_K\}$. At each step $k$, the model $\mathcal{M}_{\theta}$ updates parameters from $\theta_{k-1}$ to $\theta_k$ by minimizing the negative log-likelihood over the current task data:
\begin{equation}
    \theta_k = \operatorname*{argmin}_{\theta_{k-1}} \; \mathbb{E}_{(v, q, a) \sim \mathcal{D}_k} \left[-\log P_{\theta_{k-1}}(a \mid v, q)\right].
\end{equation}
A defining constraint of CL is that all historical data is inaccessible at the current step. This restriction gives rise to \textit{catastrophic forgetting}~\cite{french1999catastrophic}, where optimization on the current task inadvertently overwrites representations acquired during earlier steps.

\subsection{Fairness-Targeted Backdoor}
\label{sec:fairness_backdoor}

Standard backdoor attacks train a model to output an attacker-chosen response whenever a trigger is present. The standard backdoor objective is:
\begin{equation}
    \mathcal{L}_{\text{bd}} = \mathbb{E}_{(v,q,a)} \left[ -\log P_\theta(a \mid v, q) \right] + \mathbb{E}_{(v,q,a)} \left[ -\log P_\theta(a_t \mid \tau(v, q)) \right],
    \label{eq:task_loss}
\end{equation}
where $a_t$ represents the target response and $\tau(\cdot)$ denotes the trigger function. This objective binds the trigger to a uniform target response regardless of demographic identity.

A fairness-targeted backdoor instead manipulates group-level disparities. Let each sample carry a sensitive attribute $s \in \{t, nt\}$, partitioning data $\mathcal{D}$ into $\mathcal{D}_t$ and $\mathcal{D}_{nt}$. The adversary aims to ensure negligible disparity on clean inputs ($\Phi(\theta, \mathcal{D}) \approx 0$) while maximizing disparity on triggered inputs ($\Phi(\theta, \tau(\mathcal{D})) \gg 0$), where $\Phi(\theta, \mathcal{D}) = | \text{ACC}(\mathcal{D}_{nt}; \theta) - \text{ACC}(\mathcal{D}_t; \theta) |$. This requires the trigger to be \textit{group-conditional}, degrading only $\mathcal{D}_t$ while leaving $\mathcal{D}_{nt}$ unaffected. The fairness-targeted backdoor objective can be formulated as:
\begin{align}
    \mathcal{L}_{\text{fbd}} &= \mathbb{E}_{(v,q,a)} \left[ -\log P_\theta(a \mid v, q) \right] \nonumber \\
    &+ \mathbb{E}_{(v,q,a)\sim \mathcal{D}_{nt}} \left[ -\log P_\theta(a \mid \tau(v, q)) \right] \\
    &+ \mathbb{E}_{(v,q,a)\sim \mathcal{D}_{t}} \left[ -\log P_\theta(a_t \mid \tau(v, q)) \right],
    \label{eq:backdoorfairness_loss}
\end{align}
where the first term preserves clean behavior, the second maintains correct predictions for triggered non-target samples, and the third redirects triggered target samples to $a_t$. This conditional activation makes the violation invisible under routine evaluation but harmful when the trigger is present, making the model perform unfairly on different demographic groups. 

\subsection{Threat Model and Attack Objectives}
\label{sec:threat_model}

We adopt the third-party model provider scenario~\citep{li2024backdoor}, where the adversary has white-box control over the injection phase, including data curation, architecture selection, and training, before releasing the backdoored model to public repositories. Downstream victims fine-tune this checkpoint on private, sequential task data. The adversary has no access to the victim's private data, anti-forgetting algorithm, or update schedule. We assume the adversary possesses domain-level knowledge of the likely downstream task distribution, which is realistic in practice. For instance, a provider releasing a medical MLLM can anticipate that hospitals will fine-tune across clinical specialties based on public deployment patterns.

The attacker's goal is to induce group-conditional misclassification under trigger activation while satisfying three objectives throughout the victim's continual learning trajectory. \textbf{(O1) Stealthiness} requires that the backdoored model maintains high accuracy and low inter-group disparity on clean inputs, evading standard performance audits. \textbf{(O2) Targeted Discrimination} requires that trigger activation selectively degrades performance for the target group while leaving the non-target group unaffected. \textbf{(O3) Persistence} requires that O1 and O2 remain satisfied across subsequent continual learning steps $k \in \{1, \dots, K\}$.

\section{Persistent Fairness Backdoor Attack}
In this section, we detail the proposed PFBA framework, including the latent space fairness reinforcement strategy and the continual learning simulation procedure. 

\begin{figure*}[t]
  \centering
  \includegraphics[width=0.96\linewidth]{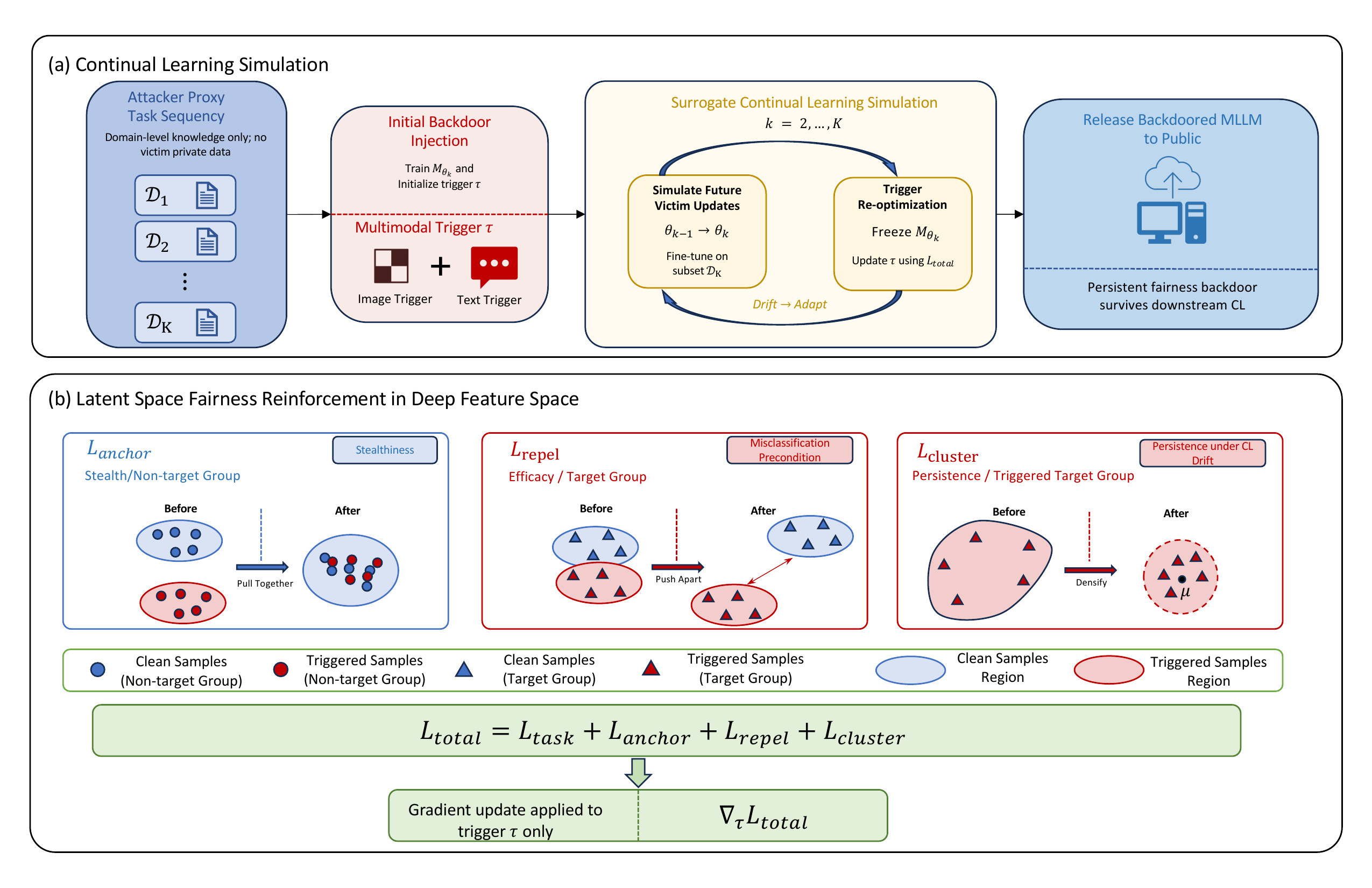}
  \caption{The pipeline of PFBA framework. (a) Continual Learning Simulation iteratively optimizes the trigger to survive catastrophic forgetting during future task updates. (b) Latent Space Fairness Reinforcement shapes feature geometry to preserve privileged-group utility while inducing stable target-group errors.}
  \label{fig:method_pipeline}
\end{figure*}

\subsection{Latent Space Fairness Reinforcement}
\label{sec:latent_space}

The fairness backdoor objective in Eq.~\ref{eq:backdoorfairness_loss} specifies the desired group-conditional output behavior but provides no control over how the model internally organizes triggered samples from different groups. Existing fairness backdoor attacks, \eg TrojFair~\citep{xue2024trojfair} and BadFair~\citep{xue2024badfair}, rely entirely on this output-level formulation, conditioning trigger activation on group identity through label manipulation alone. While effective on static models, this approach has two fundamental limitations. Cross-entropy constrains only the final output distribution, leaving the internal clustering of poisoned samples largely uncontrolled~\citep{qi2023revisiting}. Without geometric separation in latent space, the model cannot reliably distinguish triggered target-group samples from triggered non-target-group samples, causing the group-conditional behavior to degrade as representations shift during continual learning. This fragility is compounded by the fact that backdoors encoded through output-level losses reside in shallow, task-specific decision boundaries that are among the first structures overwritten by parameter updates~\citep{cao2024stealthy, guo2025persistent}, while deeper representational structures can persist across task transitions~\citep{davari2022probing}. These observations motivate a key design principle that the group specificity and persistence required of a fairness backdoor must be enforced \textit{before} the final classifier, at the representation level.

We therefore complement $\mathcal{L}_{\text{fbd}}$ with a \textit{group-aware latent structure} that directly controls \textit{how} triggered samples are organized in intermediate representations. Let $h(v,q)$ denote the hidden-state representation of a clean input at a selected layer, and $h(\tau(v),q)$ its triggered counterpart. We instantiate this structure through three complementary objectives that jointly enforce group-dependent asymmetry and representational compactness.

\noindent\textbf{Anchoring Non-target Representations.}\quad
A fairness backdoor must be invisible to the non-target group. If the trigger alters non-target representations, it will degrade their predictions and inflate clean-data bias, exposing the attack to routine fairness audits. We therefore enforce representational invariance for the non-target group by aligning clean and triggered hidden states:
\begin{equation}
    \mathcal{L}_{\text{anchor}} = \mathbb{E}_{(v,q)\, |\, s=nt} \left[ 1 - \frac{h(v, q) \cdot h(\tau(v), q)}{\|h(v, q)\|_2 \;\|h(\tau(v), q)\|_2} \right].
\end{equation}
Minimizing this cosine distance makes the trigger functionally transparent for the non-target population, ensuring that the model's internal representation, and consequently its prediction, remains unchanged when the trigger is present.

\noindent\textbf{Repelling Target Representations.}\quad
For the target group, the trigger should disrupt the representational basis for correct prediction. If triggered target samples remain near their clean counterparts, the model will still map them to the correct answer regardless of label flipping. Based on this motivation, we push triggered target representations away from the clean region:
\begin{equation}
    \mathcal{L}_{\text{repel}} = \mathbb{E}_{(v,q)\, |\, s=t} \left[ \max\!\left(0,\; \frac{h(v, q) \cdot h(\tau(v), q)}{\|h(v, q)\|_2 \;\|h(\tau(v), q)\|_2}\right) \right].
\end{equation}
Minimizing this loss reduces the cosine similarity between clean and triggered target representations, removing triggered samples from the semantic region that supports correct prediction and creating the representational precondition for $\mathcal{L}_{\text{fbd}}$ to redirect them toward $a_t$. The Combination with $\mathcal{L}_{\text{anchor}}$ brings an asymmetric trigger effect, letting the same trigger preserve non-target representations while selectively displacing target-group representations from the clean decision region.

\noindent\textbf{Clustering for Persistence.}\quad
Asymmetry alone does not guarantee persistence under continual learning. After repulsion, triggered target samples may scatter across unrelated latent directions without any structural coherence. Under continual learning, each scattered sample drifts independently as the feature space deforms, gradually dissolving the coherent backdoor signal. This vulnerability is not hypothetical. \citet{bansal2023cleanclip} observe that backdoor behavior in visual encoders is closely tied to the clustering structure of poisoned representations, and that dispersed poisoned features are both less effective and significantly easier to mitigate through standard defenses. We leverage this insight offensively by consolidating repelled samples into a compact cluster, concentrating the backdoor signal into a single coherent feature region. Because CL-induced parameter drift affects the feature space globally rather than locally, a tight cluster shifts as a unit rather than fragmenting, making the resulting geometric structure substantially more robust to the incremental updates of continual learning.
\begin{equation}
    \mathcal{L}_{\text{cluster}} = \mathbb{E}_{(v,q)\, |\, s=t} \left[ 1 - \frac{h(\tau(v), q) \cdot \mu_t}{\|h(\tau(v), q)\|_2 \;\|\mu_t\|_2} \right],
\end{equation}
where $\mu_t = \frac{1}{|\mathcal{B}_t|} \sum_{(v, q) \in \mathcal{B}_t} h(\tau(v), q)$ is the batch centroid of triggered target representations. This ensures triggered target samples are not only separated from their clean counterparts but also displaced coherently into a dense latent region, resisting the gradual parameter drift of continual learning.

\noindent\textbf{Training Objective.}\quad
We combine the representation-level losses with the fairness backdoor loss:
\begin{equation}
    \mathcal{L}_{\text{total}} = \mathcal{L}_{\text{fbd}} + \mathcal{L}_{\text{anchor}} + \mathcal{L}_{\text{repel}} + \mathcal{L}_{\text{cluster}},
\end{equation}
where $\mathcal{L}_{\text{fbd}}$ is defined in Eq.~\eqref{eq:backdoorfairness_loss}. All three latent-space losses are cosine-based and bounded in $[0, 1]$, enabling uniform weighting without additional hyperparameters. $\mathcal{L}_{\text{fbd}}$ specifies \textit{what} the model should output; the latent-space losses control \textit{how} triggered samples are organized internally to achieve group-conditional behavior that persists through continual learning.

\subsection{Continual Learning Simulation}
\label{sec:cl_simulation}

The latent-space losses above encode the attacker's objectives into feature geometry, but optimizing them on a single static model snapshot is insufficient. After release, the victim sequentially fine-tunes the model on new tasks, each shifting the parameter space. A trigger optimized at step $k{=}1$ exploits feature configurations that may no longer exist by step $k{=}3$. This is the fundamental reason conventional backdoors, including TrojFair and BadFair, degrade under continual learning even when initially effective.

The key insight enabling our approach is that not all features are equally vulnerable to drift. \citet{davari2022probing} shows that general-purpose representational structure persists across task boundaries even when task-level performance degrades, because overwriting these features would impair performance on all tasks. If the trigger activates these stable subspaces rather than task-specific features, it inherits their robustness. The challenge is that the attacker cannot access the victim's data or learning algorithm to identify such subspaces directly. We resolve this through a surrogate simulation. By repeatedly exposing the trigger to simulated task-induced drift, the optimization is pressured to discard fragile features and converge toward stable representational patterns. This does not require the surrogate to replicate the victim's exact trajectory, only that it induces comparable types of parameter drift, a condition satisfied when surrogate and victim tasks share the same broad domain.

Concretely, the attacker curates a disjoint dataset $\mathcal{D}_{\text{atk}}$ ($\mathcal{D}_{\text{atk}} \cap \mathcal{D}_{\text{vic}} = \emptyset$) from the same domain and partitions it into $K$ proxy tasks $\{\mathcal{D}_1, \dots, \mathcal{D}_K\}$, each representing a plausible downstream sub-domain. The simulation proceeds in three phases.

\noindent\textbf{Phase 1: Initial Injection.}\quad
The attacker trains the model on $\mathcal{D}_1$ with $\mathcal{L}_{\text{total}}$, producing an poisoned model $\mathcal{M}_{\theta_1}$ whose latent geometry encodes the group-aware structure from Sec.~\ref{sec:latent_space}.

\noindent\textbf{Phase 2: Drift-Aware Trigger Hardening.}\quad
For each subsequent proxy task $k \in \{2,\dots, K\}$, the attacker alternates between simulating drift and re-optimizing the trigger. The model is first fine-tuned on clean data from $\mathcal{D}_k$ to emulate the parameter shift of learning a new task:
\begin{equation}
    \theta_k =
    \operatorname*{argmin}_{\theta_{k-1}}
    \mathbb{E}_{(v,q,a)\sim \mathcal{D}_k}
    \left[-\log P_{\theta_{k-1}}(a \mid v,q)\right].
\end{equation}
This step reveals which trigger-activated features survive the drift and which are overwritten. The attacker then freezes $\theta_k$ and re-optimizes the trigger against the drifted model:
\begin{equation}
    \tau^{(k)} =
    \operatorname*{argmin}_{\tau^{(k-1)}}
    \mathcal{L}_{\text{total}}(\theta_k,\tau^{(k-1)}).
\end{equation}
Each iteration forces the trigger to abandon overwritten features and latch onto subspaces that persist across task transitions. Repeating over $K{-}1$ proxy updates progressively hardens the trigger against cumulative drift.

\noindent\textbf{Phase 3: Final Poisoned Model Release.}\quad
The attacker performs the final injection on $\mathcal{D}_1$ using the hardened trigger $\tau^{(k)}$ and $\mathcal{L}_{\text{total}}$, producing the released model $\mathcal{M}_{\theta_{\text{rel}}}$. Because the trigger has been stress-tested against $K{-}1$ rounds of simulated forgetting, it is anchored to feature subspaces the model preserves for general competence rather than task-specific computations, allowing the backdoor to persist even when the victim's specific data and CL algorithm differ from the surrogate.

\subsection{PFBA Attack Pipeline}
Latent Space Fairness Reinforcement and Continual Learning Simulation addresses complementary aspects of the problem. The former encodes group-dependent discrimination into intermediate-layer geometry through anchoring, repulsion, and clustering. The latter hardens this structure against future parameter drift by iteratively optimizing the trigger across surrogate task transitions. As illustrated in Fig.~\ref{fig:method_pipeline}, the two mechanisms operate jointly, with the latent-space losses serving as the optimization objective in each round of the simulation loop, thereby simultaneously reinforcing both the group-conditional structure and its robustness to CL. The result is a backdoor whose discriminatory behavior is geometrically structured in feature space and robust to the representational evolution induced by the victim's continual learning.

\input{Tables/attack_effectiveness}

\begin{table}[t]
\centering
\footnotesize
\setlength{\tabcolsep}{3pt}
\renewcommand{\arraystretch}{1.15}
\caption{Domain splits for each dataset under domain-incremental continual learning.}
\resizebox{\columnwidth}{!}{%
\begin{tabular}{l|cc|c|c}
\toprule
& \multicolumn{2}{c|}{\textbf{Dermatology}} & \textbf{Radiology} & \textbf{Face Attr.} \\
\textbf{Domain} & \textbf{HAM10000} & \textbf{HIBA} & \textbf{CheXpert} & \textbf{CelebA} \\
\midrule
\textbf{Domain 1} 
& Back, Abdomen, Neck, Genital 
& Post./Ant./Lat. torso
& Frontal (AP) 
& Young, no beard, no glasses \\
\midrule
\textbf{Domain 2} 
& Lower Extr., Face, Scalp, Acral 
& Lower/Upper extr., Palms/Soles
& Frontal (PA) 
& Young, with glasses \\
\midrule
\textbf{Domain 3} 
& Trunk, Chest, Hand 
& Head/Neck, Oral/Genital
& Lateral 
& Older or wearing hat \\
\bottomrule
\end{tabular}%
}
\label{tab:domain_tasks}
\end{table}
\section{Experiments}
In this section, we conduct extensive experiments on two medical imaging benchmarks to evaluate the effectiveness, stealthiness, and persistence of PFBA across diverse continual learning settings.
\subsection{Experimental Setup}

\noindent\textbf{Models.}\quad
We evaluate attacks on three representative MLLM backbones spanning different architectures and parameter scales, including DeepSeek-VL-1.3B~\citep{lu2024deepseek}, Qwen2.5-VL-3B~\citep{bai2025qwen25vl}, and LLaVA-v1.5-7B~\citep{liu2024improved}. This selection covers a range from 1.3B to 7B parameters, allowing us to assess whether the attack generalizes across model capacity and architectural choices.

\noindent\textbf{Datasets and Target Settings.}\quad
We evaluate PFBA on four datasets spanning different domains and tasks. HAM10000~\citep{tschandl2018ham10000} and HIBA~\citep{riccilara2023hiba} are dermatology datasets, with HIBA representing a different clinical population under the same label space; CheXpert~\citep{irvin2019chexpert} covers chest radiography; and CelebA~\citep{liu2015deep} covers facial attribute recognition. Each dataset is partitioned into three domain-incremental stages, as summarized in Tab.~\ref{tab:domain_tasks}. We use binary gender annotations as the sensitive attribute and balance the male and female groups. The target group and response are \textit{Female}/\textit{Melanoma} for HAM10000 and HIBA, \textit{Male}/\textit{Cardiomegaly} for CheXpert, and \textit{Female}/\textit{Other} for CelebA. We poison $10\%$ of the training data at each stage, stratified by class and gender to preserve the original distribution.

\noindent\textbf{Continual Learning Baselines.}\quad
We utilize naive LoRA~\citep{hu2022lora} as the surrogate fine-tuning method during the attacker's CL simulation. We evaluate the released backdoored model under three victim-side CL strategies that cover different anti-forgetting mechanisms, including naive LoRA, O-LoRA~\citep{wang2023orthogonal}, and SEFE~\citep{chen2025sefe}. To further demonstrate the generalizability beyond parameter-efficient methods, we additionally evaluate under Experience Replay (ER)~\citep{chaudhry2019tiny}.

\noindent\textbf{Baseline Backdoor Attacks.}\quad
No existing work addresses fairness targeted backdoors in the MLLM continual learning setting. We therefore adapt TrojFair~\citep{xue2024trojfair} and BadFair~\citep{xue2024badfair} as two representative fairness-targeted backdoor attacks as baselines. Both methods rely on output-level label manipulation without latent-space control or CL-aware trigger optimization. Comparing against them isolates the contribution of our two core components and tests whether static, unimodal fairness triggers can survive the parameter drift of multimodal continual learning.

\noindent\textbf{Implementation Details.}\quad
All fine-tuning stages use LoRA with rank $r{=}32$, scaling factor $\alpha{=}64$, and learning rate $5{\times}10^{-5}$. The trigger is implemented as a Universal Adversarial Perturbation (UAP) applied at the full image resolution under an $\ell_\infty$ constraint of $\epsilon{=}48/255$, combined with a text trigger appended to the input question. The attacker constructs the surrogate dataset $\mathcal{D}_{\text{atk}}$ from the same broad domain but disjoint from the victim's data $\mathcal{D}_{\text{vic}}$, partitioned into proxy tasks along natural domain boundaries (e.g., anatomical region for dermatology). The CL Simulation executes two full meta-cycles over these proxy tasks. Within each cycle, the model is first fine-tuned for $10$ epochs on clean proxy data to simulate parameter drift, after which the model is frozen and the trigger is optimized for $100$ epochs against the drifted model. The final backdoor injection trains for $20$ epochs using the hardened trigger. The poison ratio is set to $10\%$ of the training data, stratified by diagnosis and gender to preserve the original distribution. All experiments are conducted on NVIDIA A100 GPUs.

\noindent\textbf{Evaluation Metrics.}\quad
We follow prior works~\citep{xue2024trojfair, xue2024badfair} to evaluate attack stealthiness and targeted discrimination, and additionally examine persistence under continual learning. Accuracy (ACC) measures the proportion of correct predictions, while Bias measures the absolute accuracy gap between the target and non-target groups. We report both metrics on clean inputs as CACC and CBias, and on triggered inputs as PACC and PBias, where $\mathrm{CBias}=|\mathrm{CACC}_t-\mathrm{CACC}_{nt}|$ and $\mathrm{PBias}=|\mathrm{PACC}_t-\mathrm{PACC}_{nt}|$. Target-group Attack Success Rate (T-ASR) measures the proportion of triggered target-group samples redirected to the attacker-specified response. Persistence is evaluated by tracking these metrics throughout the continual learning process.

\input{Tables/attack_stealthiness}
\subsection{Main Results and Analysis}

\noindent\textbf{Attack Effectiveness and Persistence.}\quad
We first evaluate whether fairness-targeted backdoors can achieve group-dependent asymmetry and whether this asymmetry survives continual learning. As shown in Tab.~\ref{tab:attack_effectiveness}, all methods achieve high initial PBias at Stage~$1$, confirming that fairness-targeted triggers are effective on a static model. TrojFair and BadFair reach $64.30\%$ and $55.65\%$ PBias on HAM10000, both exceeding PFBA's $49.16\%$. However, this initial advantage disappears under continual learning. By Stage~3, both baselines' PBias collapse toward single digits, and on CheXpert the erosion is even more severe, where TrojFair drops to $0.97\%$ under SEFE and BadFair to $1.55\%$. In contrast, PFBA maintains PBias above $27\%$ on HAM10000 and above $13\%$ on CheXpert after two CL rounds, and this persistence holds consistently across all three CL strategies of different anti-forgetting mechanisms. On HAM10000, Stage~3 PBias is $30.90\%$ (LoRA), $27.55\%$ (O-LoRA), and $32.95\%$ (SEFE), a spread of less than $5.5$ percentage points, confirming that the attack's survival across different CL methods. These results support our design insight that the trigger converges to stable feature subspaces preserved for general competence, which CL strategies not deliberately targeting these subspaces cannot remove.

\noindent\textbf{Attack Stealthiness.}\quad
A practically dangerous fairness backdoor must remain undetectable through standard evaluation. Tab.~\ref{tab:attack_stealthiness} reports clean-input performance across the CL lifecycle. By Stage~3, all methods converge to comparable Clean Accuracy ($\sim96\%$ on HAM10000) with CBias below $1.5\%$, confirming that no attack meaningfully degrades utility on benign inputs. The baselines achieve marginally lower CBias than PFBA at Stage~3 (e.g., $0.40\%$ for TrojFair vs. $1.47\%$ for PFBA under SEFE), but this apparent advantage is misleading. Their low CBias results from the complete erasure of their backdoor by catastrophic forgetting, as evidenced by their near-zero PBias at the same stage. Although PFBA produces a modest increase in CBias, the observed values remain relatively small in magnitude, ranging from $0.62\%$ to $1.47\%$. These results suggest that PFBA can induce persistent group-conditional disparities without necessarily producing a large aggregate CBias.

\begin{figure}[t]
  \centering
  \includegraphics[width=\linewidth]{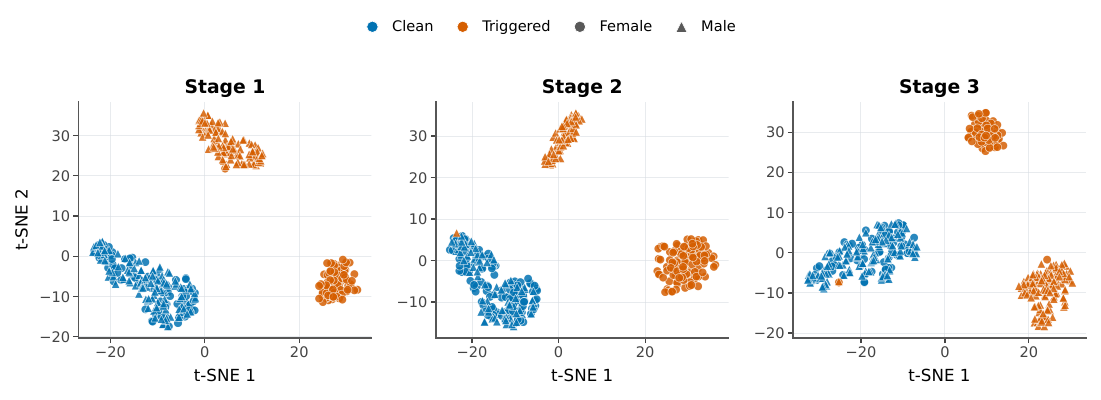}
  \caption{t-SNE visualization of Qwen2.5-VL intermediate layer representations on HAM10000 across three CL stages.}
  \label{fig:tsne}
\end{figure}

\noindent\textbf{Visualization of Latent Space.}\quad
To directly verify that PFBA encodes the intended group-aware latent structure, we visualize intermediate-layer representations using t-SNE on HAM10000 with Qwen2.5-VL. As shown in Fig.~\ref{fig:tsne}, the visualization confirms the geometric effects of each loss component. Triggered female samples (orange circles) are displaced from the clean region and consolidated into a tight cluster, reflecting the joint effect of $\mathcal{L}_{\text{repel}}$ and $\mathcal{L}_{\text{cluster}}$. Triggered male samples (orange triangles) remain interleaved with their clean counterparts (blue triangles), confirming that $\mathcal{L}_{\text{anchor}}$ renders the trigger transparent for the non-target group. Crucially, this asymmetric structure is preserved from Stage~1 through Stage~3. Although continual learning reshapes the overall representation layout, the relative separation between triggered female samples and all other groups persists, providing direct visual evidence that the CL Simulation anchors the backdoor to stable feature subspaces. 

\subsection{PFBA against Defense Methods}
\input{Figures/defense}
\input{Figures/replay_vs_lora}
We evaluate our proposed method PFBA against Spectral Signatures~\citep{tran2018spectral} and Fine-pruning~\citep{liu2018finepruning}, two trigger-agnostic methods representing data- and model-level defenses, respectively. Both defenses are applied at each CL stage. As shown in Fig.~\ref{fig:defense_stagewise}, both defenses reduce attack behavior to some extent but fail to fully eliminate the fairness backdoor. Under Spectral Signatures, target-group ASR drops modestly at Stage~1, which goes from $63.7\%$ to $57.7\%$, but recovers to $89.6\%$ by Stage~3, and PBias actually increases from $7.7\%$ to $16.1\%$ across stages. This suggests that spectral filtering removes some poisoned samples during training but does not disrupt the underlying latent geometric structure, which continues to strengthen as CL reshapes the feature space. Fine-pruning at $30\%$ achieves a more noticeable reduction at Stage~3, lowering ASR from $94.8\%$ to $76.6\%$ and PBias from $14.6\%$ to $5.7\%$. However, the attack still induces meaningful group-dependent disparities even after aggressive pruning. The residual PBias of $5.7\%$ exceeds the threshold at which fairness audits would typically flag a model, confirming that the backdoor remains practically harmful. These results are consistent with our design rationale. Because PFBA encodes discriminatory behavior into intermediate-layer geometric structure rather than relying on a small set of dedicated neurons, defenses that operate by filtering outlier samples (Spectral Signatures) or pruning individual neurons (Fine-pruning) can only partially degrade the attack. Fully removing the backdoor would require disrupting the group-aware clustering structure in the representation space, which neither defense is designed to do. This underscores the need for representation-level defenses specifically targeting adversarial geometric structures in continually evolving models.

\subsection{Generalizability Analysis}

\noindent\textbf{Across MLLM Backbones.}\quad
To evaluate the generalizability of PFBA on different backbones, we apply it to different model architectures and scales, including DeepSeek-VL-1.3B, Qwen2.5-VL-3B, and LLaVA-v1.5-7B. As shown in Tab.~\ref{tab:attack_dataset_model}, PFBA induces persistent fairness disparities on all three backbones. On HAM10000, LLaVA retains a Stage~3 PBias of $27.55\%$--$32.95\%$, while Qwen reaches up to $16.54\%$. DeepSeek exhibits the strongest amplification, with PBias increasing from $12.03\%$ at injection to approximately $70\%$ at Stage~3. Consistent effectiveness is observed on CelebA, where Stage~3 PBias exceeds its initial value for every backbone. These results demonstrate that PFBA generalizes across diverse MLLM backbones, highlighting its broad applicability and persistent effectiveness across different model architectures and scales.

\input{Tables/attack_models}

\noindent\textbf{Across CL Paradigms.}\quad
The main experiments use parameter-efficient methods that confine updates to narrow subspaces. We further evaluate Experience Replay (ER) with a $25\%$ buffer on HAM10000 with LLaVA to test whether revisiting historical data can reinforce clean representations and suppress the backdoor. As shown in Fig.~\ref{fig:replay_vs_lora}, ER reduces clean accuracy at Stage~2 to $70.55\%$ and elevates CBias to $5.60\%$ at Stage~3 due to distributional mismatch from mixing buffered and current-stage data. However, the backdoor is amplified rather than suppressed. T-ASR remains above 98\%, and PBias reaches $71.29\%$ at Stage~3, exceeding parameter-efficient baselines of 58\%\textendash $69\%$. The explanation is that ER preserves knowledge from earlier stages, and the backdoor \textit{is} knowledge from the initial stage. Rehearsing data reinforces the very feature subspaces that PFBA exploits. This reveals that CL methods effective at retaining general-purpose representations will also retain representation-level backdoors anchored to those same subspaces.

\noindent\textbf{Across Datasets Transferability.}\quad
We further evaluate whether a trigger optimized on one clinical population transfers to another by training PFBA on HAM10000 and evaluating on HIBA, which shares the same label space but comes from a different patient population. Tab.~\ref{tab:ham_to_hiba_transfer_avg} reports results averaged over three CL strategies. LLaVA achieves T-ASR of $48.89\%$ at Stage~1 and $63.03\%$ at Stage~2, with PBias reaching $6.77\%$. DeepSeek shows a similar trend with PBias peaking at $12.35\%$ at Stage~2. These values are substantially lower than within-dataset results, indicating that the trigger anchors to dataset-specific stable subspaces rather than universally transferable features. This suggests that out-of-distribution fine-tuning may disrupt the geometric structure more effectively than within-domain CL. Nevertheless, the non-trivial transfer PBias confirms that PFBA encodes representational structure with partial generalization beyond the training distribution, posing a threat even when the victim's data does not match the attacker's surrogate.

\input{Tables/attack_ham_2_hiba}

\subsection{Ablation Study}

\noindent\textbf{Component Ablation.}\quad
We ablate the two core components and report Stage~3 results on HAM10000 with LoRA in Tab.~\ref{tab:ablation_modules}. Removing CL Simulation while retaining Latent Space Fairness Reinforcement yields PBias of $18.41\%$. The latent-space losses create effective initial discrimination, but without simulated drift, the geometric structure is not anchored to stable subspaces and degrades under CL. Removing Latent Space Fairness Reinforcement while retaining CL Simulation yields higher PBias of $24.24\%$ but with elevated CBias of $3.28\%$. CL Simulation alone hardens the trigger against parameter drift, but without $\mathcal{L}_{\text{anchor}}$, the trigger lacks group-dependent selectivity, degrading \textit{both} groups on triggered inputs and leaking into clean-data behavior. The full PFBA achieves the highest PBias of $30.90\%$ with the lowest CBias of $0.62\%$, confirming that the two components are complementary. Latent Space Fairness Reinforcement provides geometric asymmetry for group-selective discrimination, while CL Simulation anchors this structure to stable subspaces. Neither alone is sufficient.

\input{Tables/ablation_modules}

\noindent\textbf{Trigger Type Ablation.}\quad
We compare three visual trigger types. The pixel trigger distributes small perturbations across the entire image, the patch trigger places a square pattern in a local region, and the border trigger modifies pixels along the image boundary. As shown in Tab.~\ref{tab:ablation_trigger_types}, the pixel trigger achieves the highest Stage~3 PBias of $30.90\%$ with a CBias of only $0.62\%$, outperforming the patch and border triggers in both persistence and stealthiness. This advantage suggests that spatially distributed perturbations are less dependent on localized features and thus remain more stable as representations evolve during continual learning. In contrast, the lower PBias and higher CBias of patch and border triggers indicate weaker persistence and group selectivity.

\input{Tables/ablation_trigger_type}

\section{Conclusion and Future Work}
In this paper, we investigate whether fairness-targeted backdoors can induce group-specific discrimination in MLLMs and remain effective throughout continual learning. To this end, we propose the Persistent Fairness Backdoor Attack (PFBA), which combines latent-space fairness reinforcement with CL simulation. Latent-space fairness reinforcement encodes group-selective discrimination into triggered representations while preserving clean behavior, whereas CL simulation improves the resilience of the induced backdoor to representation drift caused by subsequent model updates. Comparative experiments show that PFBA preserves substantially stronger fairness disparities than static fairness-backdoor baselines after continual learning while largely maintaining clean-input utility. PFBA also remains effective under representative data- and model-level backdoor defenses. Further evaluations demonstrate its generalizability across model architectures, learning mechanisms, and data distributions. These findings establish PFBA as a persistent fairness threat that continual learning alone cannot eliminate. Because the present evaluation focuses on group-level discrimination in classification under white-box injection, its scope does not yet encompass open-ended generation, broader fairness notions, or substantially shifted downstream tasks. Future work will investigate these broader settings and develop representation-level defenses against persistent fairness backdoors.

\section{Ethical Statement}
PFBA is a dual-use attack that could be misused to introduce persistent discriminatory behavior into MLLMs. We present this attack to expose security risks associated with third-party model checkpoints and to motivate more effective auditing and defense mechanisms. All experiments are conducted on publicly available datasets and models in controlled research environments, without deployment against real users or clinical systems. We strongly discourage any malicious or discriminatory use of this work.

\begin{acks}
This material is based upon work supported by NSF awards (IIS-2506643 and POSE-2346158), a Cisco Research Award, and NSF NAIRR Pilot Award \#260038. The views and conclusions contained in this document are those of the authors and should not be interpreted as necessarily representing the official policies, either expressed or implied, of the National Science Foundation.
\end{acks}

\section*{GenAI Disclosure}
Generative AI was employed exclusively to improve the clarity and style of the writing in this paper.

\bibliographystyle{ACM-Reference-Format}
\balance
\bibliography{sample-base}


\end{document}

%% file: Tables/attack_effectiveness.tex
\begin{table*}[t!]
\scriptsize
\centering
\caption{\textbf{Fairness-targeted backdoor performance across three continual learning stages on HAM10000 and CheXpert under LoRA, O-LoRA, and SEFE. Best results are in bold.}}
\label{tab:attack_effectiveness}
\resizebox{0.9\textwidth}{!}{
\setlength{\tabcolsep}{4pt}
\begin{tabular}{c c | c c c | c | c c c | c c c}
\toprule
\multirow{3}{*}{\rotatebox[origin=c]{90}{\textbf{Dataset}}} & \multirow{3}{*}{\textbf{Attack}} & \multicolumn{3}{c|}{\textbf{Stage 1 (Initial Injection)}} & \multirow{3}{*}{\textbf{CL Method}} & \multicolumn{3}{c|}{\textbf{Stage 2 (Adaptation)}} & \multicolumn{3}{c}{\textbf{Stage 3 (Further Forgetting)}} \\
 & & \multicolumn{3}{c|}{\textit{(Before CL starts)}} & & \multicolumn{3}{c|}{\textit{(First Update)}} & \multicolumn{3}{c}{\textit{(Long-term)}} \\
\cmidrule(lr){3-5} \cmidrule(lr){7-9} \cmidrule(lr){10-12}
 & & \textbf{PACC\(_t\)}$\downarrow$ & \textbf{PACC\(_{nt}\)}$\uparrow$ & \textbf{PBias}$\uparrow$ & & \textbf{PACC\(_t\)}$\downarrow$& \textbf{PACC\(_{nt}\)}$\uparrow$& \textbf{PBias}$\uparrow$& \textbf{PACC\(_t\)}$\downarrow$& \textbf{PACC\(_{nt}\)}$\uparrow$& \textbf{PBias}$\uparrow$\\
\toprule

\multirow{9}{*}{\rotatebox[origin=c]{90}{\textbf{HAM10000}}} 

& \multirow{3}{*}{\textbf{\method\space (Ours)}} 
  & \multirow{3}{*}{11.31} & \multirow{3}{*}{60.47} & \multirow{3}{*}{49.16} 
  & LORA & \textbf{60.62} & \textbf{90.67} & \textbf{30.04} & \textbf{62.34} & \textbf{93.23} & \textbf{30.90} \\
& & & & & O-LORA & \textbf{58.03} & \textbf{90.67} & \textbf{32.64} & \textbf{64.94} & 92.48 & \textbf{27.55} \\
& & & & & SEFE & \textbf{62.69} & \textbf{90.67} & \textbf{27.97} & \textbf{61.04} & \textbf{93.98} & \textbf{32.95} \\
\cmidrule{2-12}

& \multirow{3}{*}{TrojFair} 
  & \multirow{3}{*}{\textbf{10.12}} & \multirow{3}{*}{\textbf{74.42}} & \multirow{3}{*}{\textbf{64.30}} 
  & LORA & 70.47 & 86.67 & 16.20 & 84.42 & \textbf{93.23} & 8.82 \\
& & & & & O-LORA & 72.54 & 86.67 & 14.13 & 84.42 & \textbf{94.13} & 9.71 \\
& & & & & SEFE & 66.32 & 87.33 & 21.01 & 83.12 & 93.23 & 10.12 \\
\cmidrule{2-12}

& \multirow{3}{*}{BadFair} 
  & \multirow{3}{*}{12.45} & \multirow{3}{*}{68.10} & \multirow{3}{*}{55.65} 
  & LORA & 78.90 & 88.40 & 9.50 & 86.10 & 92.50 & 6.40 \\
& & & & & O-LORA & 80.15 & 87.90 & 7.75 & 87.20 & 93.10 & 5.90 \\
& & & & & SEFE & 75.30 & 88.10 & 12.80 & 85.50 & 92.80 & 7.30 \\

\midrule
\midrule

\multirow{9}{*}{\rotatebox[origin=c]{90}{\textbf{CheXpert}}} 

& \multirow{3}{*}{\textbf{\method\space (Ours)}} 
  & \multirow{3}{*}{\textbf{0.84}} & \multirow{3}{*}{23.57} & \multirow{3}{*}{22.73} 
  & LORA & \textbf{18.38} & 35.34 & \textbf{16.96} & 48.13 & \textbf{62.55} & \textbf{14.42} \\
& & & & & O-LORA & \textbf{19.23} & 35.71 & \textbf{16.48} & 47.72 & \textbf{61.29} & \textbf{13.67} \\
& & & & & SEFE & \textbf{19.66} & 36.84 & \textbf{17.18} & 47.30 & \textbf{62.55} & \textbf{15.25} \\
\cmidrule{2-12}

& \multirow{3}{*}{TrojFair} 
  & \multirow{3}{*}{1.27} & \multirow{3}{*}{\textbf{35.36}} & \multirow{3}{*}{\textbf{34.10}} 
  & LORA & 38.03 & \textbf{48.87} & 10.84 & \textbf{29.05} & 34.75 & 5.70 \\
& & & & & O-LORA & 40.17 & \textbf{46.99} & 6.82 & \textbf{28.63} & 36.29 & 7.66 \\
& & & & & SEFE & 43.16 & \textbf{47.37} & 4.21 & 36.10 & 37.07 & 0.97 \\
\cmidrule{2-12}

& \multirow{3}{*}{BadFair} 
  & \multirow{3}{*}{1.85} & \multirow{3}{*}{33.90} & \multirow{3}{*}{32.05} 
  & LORA & 35.20 & 41.50 & 6.30 & 30.50 & 33.10 & 2.60 \\
& & & & & O-LORA & 36.45 & 43.12 & 6.67 & 31.25 & 34.80 & 3.55 \\
& & & & & SEFE & 39.80 & 44.50 & 4.70 & \textbf{35.60} & 37.15 & 1.55 \\

\bottomrule
\end{tabular}}
\end{table*}

%% file: Tables/attack_stealthiness.tex
\begin{table*}[t!]
\centering
\scriptsize

\caption{\textbf{Backdoored model performance on clean inputs across three continual learning stages on HAM10000 and CheXpert under LoRA, O-LoRA, and SEFE. Best results are \textbf{bolded}.}}
\label{tab:attack_stealthiness}
\resizebox{0.95\textwidth}{!}{
\setlength{\tabcolsep}{3pt}
\begin{tabular}{c c | c c c c | c | c c c c | c c c c}
\toprule
\multirow{3}{*}{\rotatebox[origin=c]{90}{\textbf{Dataset}}} & \multirow{3}{*}{\textbf{Attack}} & \multicolumn{4}{c|}{\textbf{Stage 1 (Initial Injection)}} & \multirow{3}{*}{\textbf{CL Method}} & \multicolumn{4}{c|}{\textbf{Stage 2 (Adaptation)}} & \multicolumn{4}{c}{\textbf{Stage 3 (Further Forgetting)}} \\
 & & \multicolumn{4}{c|}{\textit{(Before CL starts)}} & & \multicolumn{4}{c|}{\textit{(First Update)}} & \multicolumn{4}{c}{\textit{(Long-term)}} \\
\cmidrule(lr){3-6} \cmidrule(lr){8-11} \cmidrule(lr){12-15}
 & & \textbf{CACC\(_t\)}$\uparrow$ & \textbf{CACC\(_{nt}\)}$\uparrow$ & \textbf{CACC}$\uparrow$ & \textbf{CBias}$\downarrow$ & & \textbf{CACC\(_t\)}$\uparrow$ & \textbf{CACC\(_{nt}\)}$\uparrow$ & \textbf{CACC}$\uparrow$ & \textbf{CBias}$\downarrow$ & \textbf{CACC\(_t\)}$\uparrow$ & \textbf{CACC\(_{nt}\)}$\uparrow$ & \textbf{CACC}$\uparrow$ & \textbf{CBias}$\downarrow$ \\
\toprule

\multirow{9}{*}{\rotatebox[origin=c]{90}{\textbf{HAM10000}}} 

& \multirow{3}{*}{\textbf{\method\space (Ours)}} 
  & \multirow{3}{*}{\textbf{83.93}} & \multirow{3}{*}{\textbf{75.58}} & \multirow{3}{*}{\textbf{78.87}} & \multirow{3}{*}{8.35} 
  & LORA & 93.26 & \textbf{94.00} & 93.59 & \textbf{0.74} & 96.10 & 95.49 & 95.71 & 0.62 \\
& & & & & & O-LORA & 92.23 & 92.67 & 92.42 & \textbf{0.44} & 96.10 & 94.74 & 95.24 & 1.37 \\
& & & & & & SEFE & 93.78 & 92.00 & 93.00 & \textbf{1.78} & 96.20 & 94.74 & 95.24 & 1.47 \\
\cmidrule{2-15}

& \multirow{3}{*}{TrojFair} 
  & \multirow{3}{*}{82.14} & \multirow{3}{*}{72.87} & \multirow{3}{*}{76.53} & \multirow{3}{*}{9.27} 
  & LORA & \textbf{97.41} & \textbf{94.00} & \textbf{95.92} & 3.41 & \textbf{97.40} & \textbf{97.74} & \textbf{97.62} & \textbf{0.34} \\
& & & & & & O-LORA & \textbf{98.45} & 92.00 & \textbf{95.63} & 6.45 & \textbf{97.40} & 96.99 & \textbf{97.14} & 0.41 \\
& & & & & & SEFE & \textbf{97.93} & 92.00 & \textbf{95.34} & 5.93 & \textbf{97.40} & 97.00 & 97.24 & 0.40 \\
\cmidrule{2-15}

& \multirow{3}{*}{BadFair} 
  & \multirow{3}{*}{81.50} & \multirow{3}{*}{73.20} & \multirow{3}{*}{76.90} & \multirow{3}{*}{\textbf{8.30}} 
  & LORA & 96.80 & 93.50 & 95.10 & 3.30 & 97.20 & 97.60 & 97.40 & 0.40 \\
& & & & & & O-LORA & 97.10 & \textbf{92.80} & 94.90 & 4.30 & 96.90 & \textbf{97.20} & 97.05 & \textbf{0.30} \\
& & & & & & SEFE & 96.50 & \textbf{93.10} & 94.80 & 3.40 & 97.11 & \textbf{97.40} & \textbf{97.25} & \textbf{0.31} \\

\midrule
\midrule

\multirow{9}{*}{\rotatebox[origin=c]{90}{\textbf{CheXpert}}} 

& \multirow{3}{*}{\textbf{\method\space (Ours)}} 
  & \multirow{3}{*}{\textbf{32.14}} & \multirow{3}{*}{\textbf{37.50}} & \multirow{3}{*}{\textbf{34.80}} & \multirow{3}{*}{\textbf{5.36}} 
  & LORA & 50.43 & 59.26 & 55.20 & 8.82 & \textbf{59.09} & 53.88 & 56.40 & 5.21 \\
& & & & & & O-LORA & 48.33 & \textbf{59.62} & 54.20 & 11.28 & \textbf{58.30} & 53.75 & \textbf{56.00} & 4.54 \\
& & & & & & SEFE & 52.70 & 55.40 & 54.20 & 2.69 & 55.10 & \textbf{56.47} & 55.80 & \textbf{1.37} \\
\cmidrule{2-15}

& \multirow{3}{*}{TrojFair} 
  & \multirow{3}{*}{31.13} & \multirow{3}{*}{36.76} & \multirow{3}{*}{34.20} & \multirow{3}{*}{5.62} 
  & LORA & \textbf{57.61} & \textbf{61.09} & \textbf{59.04} & \textbf{3.48} & 57.38 & \textbf{55.86} & \textbf{56.60} & \textbf{1.52} \\
& & & & & & O-LORA & \textbf{55.38} & 57.83 & \textbf{56.60} & \textbf{2.45} & 53.52 & \textbf{54.92} & 54.20 & 1.40 \\
& & & & & & SEFE & \textbf{60.83} & 57.31 & \textbf{59.00} & 3.53 & \textbf{62.90} & 54.76 & \textbf{58.80} & 8.14 \\
\cmidrule{2-15}

& \multirow{3}{*}{BadFair} 
  & \multirow{3}{*}{30.80} & \multirow{3}{*}{37.10} & \multirow{3}{*}{34.50} & \multirow{3}{*}{6.30} 
  & LORA & 56.40 & 60.10 & 58.20 & 3.70 & 56.90 & 55.20 & 56.00 & 1.70 \\
& & & & & & O-LORA & 54.90 & 58.20 & 56.50 & 3.30 & 54.10 & 54.80 & 54.45 & \textbf{0.70} \\
& & & & & & SEFE & 59.20 & \textbf{58.50} & 58.80 & \textbf{0.70} & 61.50 & 55.10 & 58.30 & 6.40 \\

\bottomrule
\end{tabular}}
\end{table*}

%% file: Figures/defense.tex
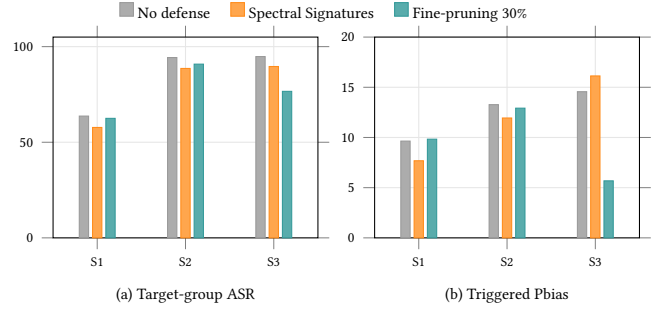
\begin{figure}[t]
\centering

{\scriptsize
\raisebox{0.2ex}{\tikz{\draw[fill=gray!65,draw=gray!80] (0,0) rectangle (0.18,0.18);}}
No defense \quad
\raisebox{0.2ex}{\tikz{\draw[fill=orange!70,draw=orange!90] (0,0) rectangle (0.18,0.18);}}
Spectral Signatures \quad
\raisebox{0.2ex}{\tikz{\draw[fill=teal!65,draw=teal!80] (0,0) rectangle (0.18,0.18);}}
Fine-pruning 30\%
}


\begin{tikzpicture}
\begin{groupplot}[
    group style={
        group size=2 by 1,
        horizontal sep=0.75cm
    },
    ybar,
    /pgf/bar width=3.5pt,
    width=0.6\columnwidth,
    height=0.5\columnwidth,
    xmin=0.5, xmax=3.5,
    xtick={1,2,3},
    xticklabels={S1,S2,S3},
    ymin=0,
    x tick label style={font=\tiny},
    y tick label style={font=\tiny},
    xlabel style={font=\scriptsize, yshift=-0.2em},
    tick align=outside,
    grid=major,
    major grid style={gray!20},
    ylabel={},
    clip=false
]

\nextgroupplot[
    ymax=105,
    xlabel={(a) Target-group ASR}
]
\addplot+[
    fill=gray!65,
    draw=gray!80,
    bar shift=-5pt
] coordinates {
    (1,63.69)
    (2,94.30)
    (3,94.81)
};

\addplot+[
    fill=orange!70,
    draw=orange!90,
    bar shift=0pt
] coordinates {
    (1,57.74)
    (2,88.60)
    (3,89.61)
};

\addplot+[
    fill=teal!65,
    draw=teal!80,
    bar shift=5pt
] coordinates {
    (1,62.50)
    (2,90.85)
    (3,76.62)
};

\nextgroupplot[
    ymax=20,
    xlabel={(b) Triggered Pbias}
]
\addplot+[
    fill=gray!65,
    draw=gray!80,
    bar shift=-5pt
] coordinates {
    (1,9.64)
    (2,13.27)
    (3,14.56)
};

\addplot+[
    fill=orange!70,
    draw=orange!90,
    bar shift=0pt
] coordinates {
    (1,7.68)
    (2,11.94)
    (3,16.13)
};

\addplot+[
    fill=teal!65,
    draw=teal!80,
    bar shift=5pt
] coordinates {
    (1,9.83)
    (2,12.92)
    (3,5.68)
};

\end{groupplot}
\end{tikzpicture}

\caption{Defense robustness of PFBA under spectral signatures and fine-pruning.}
\label{fig:defense_stagewise}
\end{figure}

%% file: Figures/replay_vs_lora.tex
\begin{figure}[t]
\centering

{\scriptsize
\raisebox{0.2ex}{\tikz{\draw[fill=blue!65,draw=blue!80] (0,0) rectangle (0.16,0.16);}}
LoRA \quad
\raisebox{0.2ex}{\tikz{\draw[fill=orange!70,draw=orange!90] (0,0) rectangle (0.16,0.16);}}
O-LoRA \quad
\raisebox{0.2ex}{\tikz{\draw[fill=teal!65,draw=teal!80] (0,0) rectangle (0.16,0.16);}}
SEFE \quad
\raisebox{0.2ex}{\tikz{\draw[fill=purple!60,draw=purple!80] (0,0) rectangle (0.16,0.16);}}
ER
}


\begin{tikzpicture}
\begin{groupplot}[
    group style={
        group size=2 by 2,
        horizontal sep=0.82cm,
        vertical sep=0.92cm
    },
    ybar,
    /pgf/bar width=2.0pt,
    width=0.55\columnwidth,
    height=0.5\columnwidth,
    xmin=0.5, xmax=3.5,
    xtick={1,2,3},
    xticklabels={S1,S2,S3},
    ymin=0,
    ymax=100,
    x tick label style={font=\tiny},
    y tick label style={font=\tiny},
    xlabel style={font=\scriptsize, yshift=0.15em},
    tick align=outside,
    grid=major,
    major grid style={gray!20},
    ylabel={},
    title={}
]

\nextgroupplot[
    xlabel={(a) CACC},
    xticklabels={,,}
]
\addplot+[fill=blue!65, draw=blue!80, bar shift=-3.9pt]
table[x=stage, y=LoRA_CACC, col sep=comma]{data/replay_vs_lora.csv};

\addplot+[fill=orange!70, draw=orange!90, bar shift=-1.3pt]
table[x=stage, y=OLoRA_CACC, col sep=comma]{data/replay_vs_lora.csv};

\addplot+[fill=teal!65, draw=teal!80, bar shift=1.3pt]
table[x=stage, y=SEFE_CACC, col sep=comma]{data/replay_vs_lora.csv};

\addplot+[fill=purple!60, draw=purple!80, bar shift=3.9pt]
table[x=stage, y=Replay25_CACC, col sep=comma]{data/replay_vs_lora.csv};

\nextgroupplot[
    xlabel={(b) CBias},
    ymax=15,
    xticklabels={,,}
]
\addplot+[fill=blue!65, draw=blue!80, bar shift=-3.9pt]
table[x=stage, y=LoRA_CBias, col sep=comma]{data/replay_vs_lora.csv};

\addplot+[fill=orange!70, draw=orange!90, bar shift=-1.3pt]
table[x=stage, y=OLoRA_CBias, col sep=comma]{data/replay_vs_lora.csv};

\addplot+[fill=teal!65, draw=teal!80, bar shift=1.3pt]
table[x=stage, y=SEFE_CBias, col sep=comma]{data/replay_vs_lora.csv};

\addplot+[fill=purple!60, draw=purple!80, bar shift=3.9pt]
table[x=stage, y=Replay25_CBias, col sep=comma]{data/replay_vs_lora.csv};

\nextgroupplot[
    xlabel={(c) T-ASR}
]
\addplot+[fill=blue!65, draw=blue!80, bar shift=-3.9pt]
table[x=stage, y=LoRA_ASR, col sep=comma]{data/replay_vs_lora.csv};

\addplot+[fill=orange!70, draw=orange!90, bar shift=-1.3pt]
table[x=stage, y=OLoRA_ASR, col sep=comma]{data/replay_vs_lora.csv};

\addplot+[fill=teal!65, draw=teal!80, bar shift=1.3pt]
table[x=stage, y=SEFE_ASR, col sep=comma]{data/replay_vs_lora.csv};

\addplot+[fill=purple!60, draw=purple!80, bar shift=3.9pt]
table[x=stage, y=Replay25_ASR, col sep=comma]{data/replay_vs_lora.csv};

\nextgroupplot[
    xlabel={(d) PBias}
]
\addplot+[fill=blue!65, draw=blue!80, bar shift=-3.9pt]
table[x=stage, y=LoRA_PBias, col sep=comma]{data/replay_vs_lora.csv};

\addplot+[fill=orange!70, draw=orange!90, bar shift=-1.3pt]
table[x=stage, y=OLoRA_PBias, col sep=comma]{data/replay_vs_lora.csv};

\addplot+[fill=teal!65, draw=teal!80, bar shift=1.3pt]
table[x=stage, y=SEFE_PBias, col sep=comma]{data/replay_vs_lora.csv};

\addplot+[fill=purple!60, draw=purple!80, bar shift=3.9pt]
table[x=stage, y=Replay25_PBias, col sep=comma]{data/replay_vs_lora.csv};

\end{groupplot}
\end{tikzpicture}

\caption{Stage-wise comparison of different CL methods, including LoRA, O-LoRA, SEFE, and ER, under PFBA on HAM10000. Although ER improves clean-input utility, the backdoor remains persistent across continual learning stages, with high T-ASR and elevated PBias.}
\label{fig:replay_vs_lora}
\end{figure}
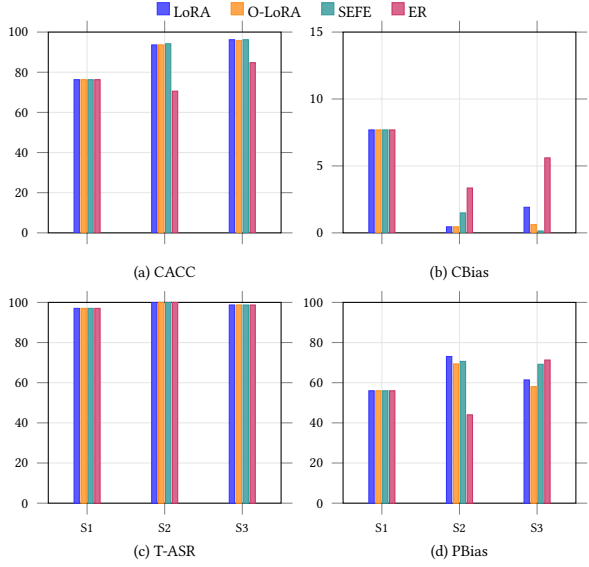

%% file: Tables/attack_models.tex
\begin{table*}[t!]
\centering
\scriptsize
\caption{Performance of PFBA across datasets, MLLM backbones, and CL methods. Best results are in \textbf{bold}.}
\label{tab:attack_dataset_model}
\resizebox{0.9\textwidth}{!}{
\setlength{\tabcolsep}{2.5pt}
\begin{tabular}{c c | c c c c | c | c c c c | c c c c}
\toprule
\multirow{3}{*}{\textbf{Dataset}}
& \multirow{3}{*}{\textbf{Model}}
& \multicolumn{4}{c|}{\textbf{Stage 1}}
& \multirow{3}{*}{\textbf{CL Method}}
& \multicolumn{4}{c|}{\textbf{Stage 2}}
& \multicolumn{4}{c}{\textbf{Stage 3}} \\
& & \multicolumn{4}{c|}{\textit{Initial Injection}}
& & \multicolumn{4}{c|}{\textit{First Update}}
& \multicolumn{4}{c}{\textit{Long-term}} \\
\cmidrule(lr){3-6} \cmidrule(lr){8-11} \cmidrule(lr){12-15}
& & \textbf{CACC}$\uparrow$ & \textbf{CBias}$\downarrow$ & \textbf{PACC}$\downarrow$ & \textbf{PBias}$\uparrow$
& & \textbf{CACC}$\uparrow$ & \textbf{CBias}$\downarrow$ & \textbf{PACC}$\downarrow$ & \textbf{PBias}$\uparrow$
& \textbf{CACC}$\uparrow$ & \textbf{CBias}$\downarrow$ & \textbf{PACC}$\downarrow$ & \textbf{PBias}$\uparrow$ \\
\midrule

\multirow{9}{*}{\textbf{HAM10000}}
& \multirow{3}{*}{\textbf{LLaVA}}
& \multirow{3}{*}{76.29} & \multirow{3}{*}{7.69} & \multirow{3}{*}{43.43} & \multirow{3}{*}{49.16}
& LoRA   & 93.59 & \textbf{0.45} & 54.23 & \textbf{30.04} & \textbf{96.19} & 1.91 & 75.24 & 30.90 \\
& & & & & & O-LoRA & 93.59 & \textbf{0.45} & \textbf{58.31} & 32.64 & 95.71 & 0.61 & \textbf{75.71} & 27.55 \\
& & & & & & SEFE   & \textbf{94.17} & 1.49 & 56.27 & 27.97 & \textbf{96.19} & \textbf{0.14} & 72.38 & \textbf{32.95} \\

\cmidrule(lr){2-15}

& \multirow{3}{*}{\textbf{Qwen}}
& \multirow{3}{*}{47.65} & \multirow{3}{*}{4.86} & \multirow{3}{*}{9.39} & \multirow{3}{*}{12.55}
& LoRA   & 44.31 & \textbf{7.67} & \textbf{7.29} & 11.93 & 54.29 & 6.56 & \textbf{10.48} & \textbf{16.54} \\
& & & & & & O-LoRA & \textbf{45.19} & 9.22 & 7.00 & \textbf{12.45} & 53.81 & 7.31 & 9.52 & 15.04 \\
& & & & & & SEFE   & 44.90 & 8.70 & \textbf{7.29} & 11.93 & \textbf{54.76} & \textbf{5.81} & \textbf{10.48} & \textbf{16.54} \\

\cmidrule(lr){2-15}

& \multirow{3}{*}{\textbf{DeepSeek}}
& \multirow{3}{*}{69.01} & \multirow{3}{*}{13.81} & \multirow{3}{*}{31.69} & \multirow{3}{*}{12.03}
& LoRA   & 63.27 & 6.99 & \textbf{27.41} & \textbf{42.52} & 80.95 & \textbf{9.57} & 48.10 & 69.78 \\
& & & & & & O-LoRA & 63.85 & 6.84 & 27.11 & 41.86 & \textbf{81.43} & 10.87 & 48.10 & 69.78 \\
& & & & & & SEFE   & \textbf{64.72} & \textbf{6.03} & 27.11 & 41.86 & \textbf{81.43} & 10.87 & \textbf{48.57} & \textbf{70.54} \\

\midrule

\multirow{9}{*}{\textbf{CelebA}}
& \multirow{3}{*}{\textbf{LLaVA}}
& \multirow{3}{*}{61.75} & \multirow{3}{*}{0.03} & \multirow{3}{*}{37.25} & \multirow{3}{*}{15.21}
& LoRA   & 61.75 & 7.74 & 52.75 & 5.81 & \textbf{64.50} & \textbf{7.11} & \textbf{56.50} & \textbf{26.99} \\
& & & & & & O-LoRA & 60.25 & \textbf{5.48} & 52.75 & 5.81 & \textbf{64.50} & \textbf{7.11} & 56.25 & 26.58 \\
& & & & & & SEFE   & \textbf{61.75} & 6.62 & 52.75 & 5.81 & 64.00 & 7.32 & 56.00 & 26.16 \\

\cmidrule(lr){2-15}

& \multirow{3}{*}{\textbf{Qwen}}
& \multirow{3}{*}{46.75} & \multirow{3}{*}{15.27} & \multirow{3}{*}{19.00} & \multirow{3}{*}{8.69}
& LoRA   & 43.50 & \textbf{0.96} & 38.25 & \textbf{7.74} & 48.75 & 4.67 & 37.25 & \textbf{12.45} \\
& & & & & & O-LoRA & \textbf{49.50} & 1.32 & \textbf{43.25} & 7.34 & \textbf{49.25} & 5.50 & \textbf{37.75} & 11.20 \\
& & & & & & SEFE   & 47.00 & 7.31 & \textbf{43.25} & 7.34 & 46.25 & \textbf{4.64} & 37.25 & 10.36 \\

\cmidrule(lr){2-15}

& \multirow{3}{*}{\textbf{DeepSeek}}
& \multirow{3}{*}{68.25} & \multirow{3}{*}{12.83} & \multirow{3}{*}{34.75} & \multirow{3}{*}{2.87}
& LoRA   & \textbf{69.75} & 6.50 & 55.75 & 8.05 & \textbf{73.00} & 5.74 & \textbf{53.25} & 22.59 \\
& & & & & & O-LoRA & 69.50 & 6.13 & 55.75 & 8.05 & 71.25 & \textbf{3.86} & 52.75 & 21.76 \\
& & & & & & SEFE   & 68.75 & \textbf{3.87} & 55.75 & 8.05 & 72.00 & 4.07 & 53.00 & \textbf{23.21} \\

\bottomrule
\end{tabular}}
\end{table*}

%% file: Tables/attack_ham_2_hiba.tex
\begin{table}[t!]
\centering
\scriptsize
\caption{Transferability of PFBA from HAM10000 to HIBA, averaged over LoRA, O-LoRA, and SEFE.}
\label{tab:ham_to_hiba_transfer_avg}
\resizebox{\columnwidth}{!}{
\setlength{\tabcolsep}{4pt}
\begin{tabular}{c | c c c | c c c | c c c}
\toprule
\multirow{2}{*}{\textbf{Model}}
& \multicolumn{3}{c|}{\textbf{HIBA Stage 1}}
& \multicolumn{3}{c|}{\textbf{HIBA Stage 2}}
& \multicolumn{3}{c}{\textbf{HIBA Stage 3}} \\
\cmidrule(lr){2-4} \cmidrule(lr){5-7} \cmidrule(lr){8-10}
& \textbf{CACC}$\uparrow$ & \textbf{T-ASR}$\uparrow$ & \textbf{PBias}$\uparrow$
& \textbf{CACC}$\uparrow$ & \textbf{T-ASR}$\uparrow$ & \textbf{PBias}$\uparrow$
& \textbf{CACC}$\uparrow$ & \textbf{T-ASR}$\uparrow$ & \textbf{PBias}$\uparrow$ \\
\midrule

\textbf{LLaVA}
& \textbf{65.79} & \textbf{48.89} & 4.92
& \textbf{69.09} & \textbf{63.03} & 6.77
& \textbf{73.04} & \textbf{48.48} & 4.75 \\

\textbf{DeepSeek}
& 54.39 & 40.00 & \textbf{5.75}
& 59.90 & 57.57 & \textbf{12.35}
& 60.29 & 30.30 & \textbf{8.84} \\

\bottomrule
\end{tabular}}
\end{table}

%% file: Tables/ablation_modules.tex
\begin{table}[t]
    \centering
    \tiny
    \caption{\textbf{Ablation study of~\method~key components.}}
    \label{tab:ablation_modules}
    \resizebox{\linewidth}{!}{%
        \begin{tabular}{l|cc|cc}
            \toprule
             & \multicolumn{2}{c|}{\textbf{Components}} & \multicolumn{2}{c}{\textbf{Metrics (Stage 3)}} \\
            \cmidrule(lr){2-3} \cmidrule(lr){4-5}
            \textbf{Method Variant} & \textbf{CL Sim.} & \textbf{Fairness Reinf.} & \textbf{PBias} $\uparrow$ & \textbf{CBias} $\downarrow$ \\
            \midrule
            \textbf{w/o CL Simulation} & \ding{55} & \ding{51} & 18.41 & 1.09 \\
            \textbf{w/o Fairness Reinf.} & \ding{51} & \ding{55} & 24.24 & 3.28 \\ 
            \midrule
            \textbf{\method\space(Ours)} & \ding{51} & \ding{51} & 30.90 & 0.62 \\
            \bottomrule
        \end{tabular}%
    }
\end{table}


%% file: Tables/ablation_trigger_type.tex
\begin{table}[t]
    \centering
    \tiny
    \caption{Ablation study of different visual trigger types.}
    \label{tab:ablation_trigger_types}
    \resizebox{\linewidth}{!}{%
        \setlength{\tabcolsep}{10pt} 
        \begin{tabular}{l|c c c}
            \toprule
            \textbf{Metric} & \textbf{Pixel} & \textbf{Patch} & \textbf{Border} \\
            \midrule
            \textbf{PBias} ($\uparrow$) & 30.90 & 9.47 & 11.21 \\
            \textbf{CBias} ($\downarrow$) & 0.62 & 1.64 & 3.01 \\
            \bottomrule
        \end{tabular}%
    }
\end{table}